\pdfoutput=1

\documentclass[11pt]{article}

\usepackage[final]{acl}

\usepackage{times}
\usepackage{latexsym}
\usepackage[T1]{fontenc}

\usepackage[utf8]{inputenc}

\usepackage{microtype}
\usepackage[noend]{algpseudocode}
\usepackage{algorithmicx,algorithm}
\usepackage{graphicx}
\usepackage{amsmath}
\usepackage{booktabs}
\usepackage{booktabs}
\usepackage{enumitem}
\usepackage{float} 
\usepackage{subfigure}
\usepackage{setspace}
\usepackage{helvet}  
\usepackage{courier}  
\usepackage{xcolor}
\usepackage{color}
\usepackage{multicol}
\usepackage{multirow}
\usepackage{esvect}
\usepackage{epstopdf}
\usepackage{array}
\usepackage{amsmath,amsfonts,amssymb,amsthm, bm}
\usepackage{mathrsfs}
\usepackage{verbatim}
\usepackage{enumitem}
\usepackage{latexsym}
\usepackage{setspace}
\usepackage{amsmath}
\usepackage{amssymb}
\usepackage{mathrsfs}

\title{ Fine- and Coarse-Granularity Hybrid Self-Attention for Efficient BERT}

\author{Jing Zhao, Yifan Wang, Junwei Bao\thanks{~~Corresponding author}, Youzheng Wu, Xiaodong He \\
JD AI Research, Beijing, China \\
\tt{\{zhaojing857,wangyifan15,baojunwei}, \\
\tt{\ wuyouzheng1,xiaodong.he\}@jd.com}}

\begin{document}

\maketitle
\begin{abstract}
Transformer-based pre-trained models, such as BERT, have shown extraordinary success in achieving state-of-the-art results in many natural language processing applications.
However, deploying these models can be prohibitively costly, as
the standard self-attention mechanism of the Transformer suffers from quadratic computational cost in the
input sequence length.
To confront this, we propose FCA, a fine- and coarse-granularity hybrid self-attention that reduces the computation cost through progressively shortening the computational sequence length in self-attention.
Specifically, FCA conducts an attention-based scoring strategy to determine the informativeness of tokens at each layer.
Then, the informative tokens serve as the fine-granularity computing units in self-attention 
and the uninformative tokens are replaced with one or several clusters as the coarse-granularity computing units in self-attention.
Experiments on GLUE and RACE datasets show that
BERT with FCA achieves 2x reduction in FLOPs over original  BERT with <1\% loss in accuracy.
We show that FCA offers significantly better
trade-off between accuracy and FLOPs
compared to prior methods\footnote{Code is available at \url{https://github.com/pierre-zhao/FCA-BERT}}.
\end{abstract}

\section{Introduction}
Transformer-based large pre-trained language models with BERT~\cite{BERT} as a typical model routinely achieve state-of-the-art results on a number of natural language processing tasks~\cite{xlnet,roberta,electra}, such as sentence classification~\cite{glue}, question answering~\cite{squad1,squad2}, and information extraction~\cite{unified}. 

Despite notable gains in accuracy, the high computational cost of these large models slows down their inference speed, which severely impairs their practicality, especially in the case of limited industry time and resources, such as Mobile Phone and AIoT. 
In addition, the excessive energy consumption and environmental impact caused by the computation of these models also raise the widespread concern~\cite{energy,green}.



To improve the efficiency of BERT, the mainstream techniques are knowledge distillation~\cite{distilling} and pruning.
Knowledge distillation aims to transfer the “knowledge"
from a large teacher model to a lightweight student model. The student model is then used during inference, such as  DistilBERT~\cite{distilbert}.
Pruning technique includes: (1) structured methods that prune structured blocks of weights or even complete architectural components in BERT, for example encoder layers~\cite{layerdrop}, (2) unstructured methods that dynamically drop redundant units, for example, attention head~\cite{head} and attention tokens~\cite{powerbert}. 
However, both types of methods encounter challenges. 
For the former, a great distillation effect often requires an additional large teacher model and very complicated training steps~\cite{tinybert,dynabert}. 
For the latter, pruning methods discard some computing units, which inevitably causes information loss.

In contrast to the prior approaches, we propose a self-motivated and information-retained technique, namely FCA, a fine- and  coarse-granularity hybrid self-attention that reduces the cost of BERT through progressively shortening the computational sequence length in self-attention.
Specifically,  FCA first evolves an attention-based scoring strategy to assign each token with the informativeness.  Through analyzing the informativeness distribution at each layer, we conclude that maintaining full-length token-level representations is progressive redundant along with layers, especially for the classification tasks that only require single-vector representations of sequences.
Consequently, the tokens are divided into informative tokens and uninformative tokens according to their informativeness.
Then, they are updated through different computation paths. 
The informative tokens carry most of the learned features and remain unchanged as the fine-grained computing units in self-attention.
The uninformative tokens may not be as important as informative ones but we will not completely discard them to avoid information loss. 
Instead, We replace them with more efficient computing units to save memory consumption.
Experiments on the standard GLUE benchmark show that FCA accelerates BERT inference speed and maintains high accuracy as well.

Our contributions are summarized as follows:
\begin{itemize}

\item We analyze the progressive redundancies in maintaining full-length token-level representations for the classification tasks.
\item We propose a  fine- and  coarse-granularity hybrid self-attention, which is able to reduce the cost of BERT and maintain high accuracy.
\item Experiments on the standard GLUE benchmark show that the FCA-based BERT achieves 2x reduction in FLOPs over the standard BERT with < 1\% loss in accuracy.

\end{itemize}

\section{Related work}
There has been much prior literature on improving the efficiency of Transformers. The most common technologies include:

\noindent \textbf{Knowledge distillation} refers to training a smaller student model  using outputs from various intermediate representations of larger pre-trained teacher models. 
In the BERT model, there are multiple representations that the student can learn from, such
as the logits in the final layer, the outputs of the
encoder units, and the attention maps.
The distillation on output logits is most commonly used to train smaller BERT models~\cite{distilbert,patient,tinybert,mobilebert}.
The output tensors of encoder units contain meaningful semantic and contextual relationships between input tokens. Some work creates a
smaller model by learning from the outputs of teacher's encoder~\cite{tinybert,mobilebert,bertemd}. 
Attention map refers to the softmax distribution output of
the self-attention layers and indicates the contextual dependence between the input tokens. A common practice of distillation on attention maps is to directly minimize the difference between the self-attention outputs of the teacher and the student~\cite{tinybert,mobilebert,ladabert}. This line of work is orthogonal to our approach and our proposed FCA can be applied to the distillate models to further accelerate their inference speed.

\noindent \textbf{Pruning} refers to identifying and removing less important weights or computation units.
Pruning methods for BERT broadly fall into two categories.
Unstructured pruning methods prune individual weights by comparing their absolute values or gradients with a pre-defined threshold~\cite{ladabert,compressing,lottery}. The weights lower than the threshold are set to zero.
Unlike unstructured pruning, structured pruning aims to prune
structured blocks of weights or even complete architectural components in the BERT model. 
\citet{head} pruned attention heads using a method based on stochastic gates and a differentiable relaxation of the L0 penalty.
\citet{reducing} randomly dropped Transformer layers to sample small sub-networks from the larger model during training which are selected as the inference models.
\citet{powerbert} progressively reduced sequence length by pruning word-vectors based on the attention values. This work is partly similar to the fine-grained computing units in our proposed FCA. However they ignored the coarse-grained units that may cause information loss.

In addition, there are some engineering techniques to speed up the inference speed, such as Mixed Precision~\cite{mixed} and Quantization~\cite{q8bert,fan}. Using half-precision or mixed-precision representations of floating points is popular in deep learning to accelerate training and inference speed. Quantization refers to reducing the number of unique values required to represent the model weights, which in turn allows to represent them using fewer bits.
\begin{figure*}[tb]
	\centering
	\includegraphics[width=16.5cm]{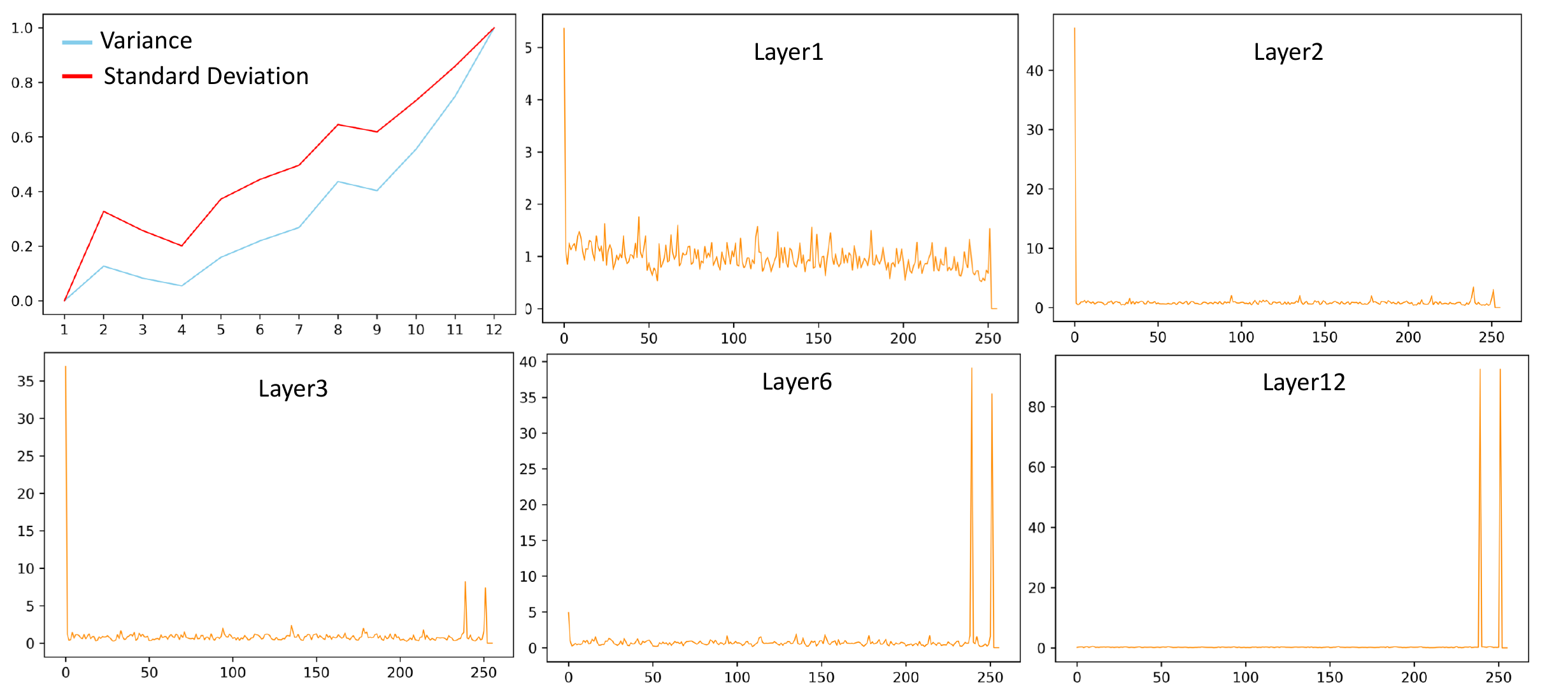}
	\caption{The first sub-figure is the normalized variance and standard deviation of informativeness with respect to BERT-base layers from 1 to 12. The last
	 five sub-figures are the informativeness distributions on some layers.}
	\label{attention_fig}
\end{figure*}

\section{Preliminary}
BERT~\cite{BERT} is a Transformer-based language representation model, which
can be fine-tuned for many downstream NLP tasks, including sequence-level and token-level classification.
The Transformer architecture~\cite{Transformer} is a highly modularized neural network, where each Transformer layer consists of two sub-modules, namely the multi-head self-attention sub-layer (MHA) and the position-wise feed-forward
network sub-layer (FFN). Both sub-modules are wrapped by a residual connection and layer normalization. 

\noindent \textbf{MHA}. The self-attention mechanism allows the model to identify complex dependencies between the elements of each input sequence. It
can be formulated as querying a dictionary
with key-value pairs. Formally,
\begin{equation}
    \text{MHA}(Q,K,V) = \text{Concat}(\text{head}_1,...,\text{head}_h)W^O
\end{equation}
where $Q,K$, and $V$ represent query, key, and
value. 
$h$ is the number of heads. 
Each head is defined as:
\begin{equation}
\begin{aligned}
    \text{head}_t &= \text{Attention}(QW_t^Q,KW_t^K,VW_t^V)  \\
    &=\underbrace{\text{softmax}(\frac{QW_t^Q(KW_t^K)^T}{\sqrt{d_K}})}_{A}VW_t^V
\label{attention}    
\end{aligned}
\end{equation}
where $W_t^Q \in \mathbb{R}^{d_h \times d_Q}, W_t^K \in \mathbb{R}^{d_h \times d_K}, W_t^V \in \mathbb{R}^{d_h \times d_V}, W^O \in \mathbb{R}^{hd_V \times d_h}$ are learned parameters. $d_K, d_Q, $and $d_V$ are dimensions of the hidden vectors. The main cost of MHA layer is the calculation of attention mapping matrix $A \in \mathbb{R}^{n \times n}$ in Eq.~\ref{attention} which is $O(n^2)$ in time and space complexity. This quadratic dependency on the sequence length has become a bottleneck for Transformers.

\noindent \textbf{FFN}. The self-attention sub-layer in each of the layers is followed by a fully
connected position-wise feed-forward network, which
consists of two linear transformations with a GeLU~\cite{gelu} activation in between. Given a vector $x_i$ in $[x_1,...,x_n]$ outputted by MHA sub-layer, FFN is defined as:
\begin{equation}
    \text{FFN}(x_i) = \text{GeLU}(x_iW_1+b_1)W_2+b_2,
\end{equation}
where $W_1,W_2,b_1,b_2$ are learned parameters.

Previous research~\cite{case} has shown that in addition to MHA sub-layer, FFN sub-layer also consumes large memory in terms of model size and FLOPs.
As a result, if we reduce the computational sequence length of MHA, the input and the consumption of FFN sub-layer will become less accordingly.

\begin{figure*}[tb]
	\centering
	\includegraphics[width=14cm]{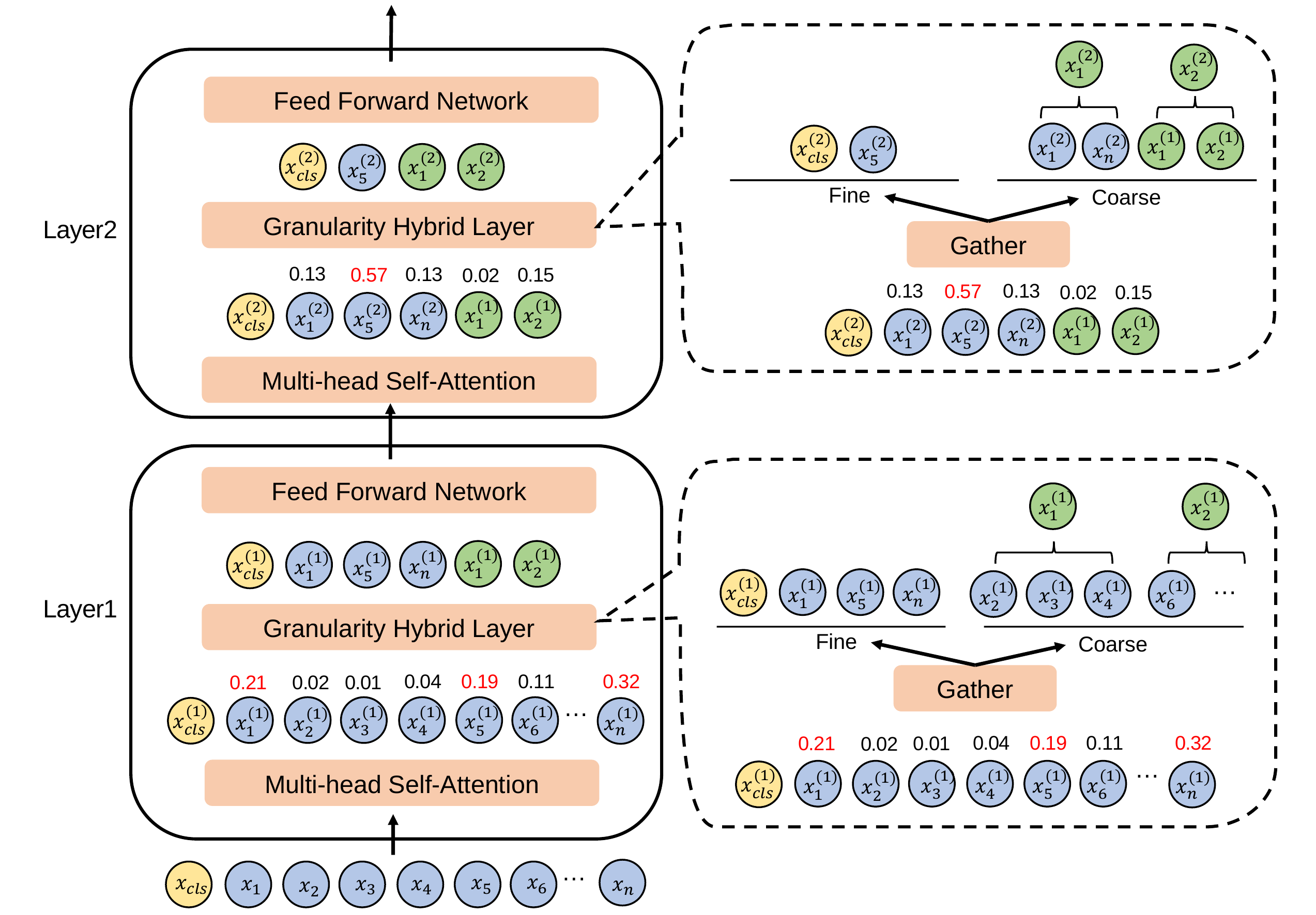}
	\caption{The architecture of FCA. The number marked above the tokens is its corresponding informativeness. The blue tokens are fine units and the green are coarse units. In this figure, we fix the number of coarse units to 2.}
	\label{overview}
\end{figure*}

\section{Methodologies}
To shorten the computational sequence length of self-attention, our core motivation is to divide tokens into informative and uninformative ones and replace the uninformative tokens with more efficient units. This section introduces each module of our model in detail.
\subsection{Scoring Strategy} 
Our strategy of scoring the informativeness of tokens is based on the self-attention map.
Concretely, taking a single token vector $x_i$ as an example, its attention head $x_i^{(t)}$ is updated by: $x_i^{(t)}=\sum_{j=1}^na_{i,j}x_j^{(t)}$ (Eq.~\ref{attention}). $a_{i,j}$ is an element in attention map $A$. Therefore, $a_{i,j}$ represents the information contribution from token vector $x_j$ to $x_i$ over $\text{head}_t$. Intuitively, we define the informativeness of a token by accumulating along the columns of attention map $A$:
\begin{equation}
    \text{I}(x_j^{(t)}) = \sum_{i=1,i\ne j}^na_{i,j}
\end{equation}

The  overall informativeness of $x_j$ is defined as the average over the heads:
\begin{equation}
    \text{I}(x_j) = \frac{1}{h}\sum_{t=1}^h\text{I}(x_j^{(t)})
\end{equation}

We next analyze some properties of defined informativeness in BERT-base. The first sub-figure in Figure~\ref{attention_fig} displays the normalized variance and standard deviation of informativeness of layers from 1 to 12 on RTE (classification dataset), which supports the phenomenon that the informativeness distributions at the bottom layers are relatively uniform and the top layers are volatile. 
The last five sub-figures further present the informativeness distributions of some BERT-base layers, where the first token is [CLS] and its representations are used for the final prediction. We can see that as the layers deepen, the informativeness is progressively concentrated on two tokens. 
This means that maintaining full-length token-level representations for the classification tasks may be redundant.

A straightforward approach for reducing the sequence length of self-attention is to maintain the informative tokens and prune the rest. We argue that this approach is effortless but encounters the risk of information loss, especially for lower layers.


\subsection{FCA Layer} 
Instead of pruning, we propose to process the uninformative tokens with more efficient units. Figure~\ref{overview} shows the architecture of FCA layer, which inserts a granularity hybrid sub-layer after MHA.
At each layer, it first divides tokens into informative and uninformative ones based on their assigned informativeness. The CLS token is always divided into informative part as it is used to derive the final prediction.

Let $x_{cls}^{(l)} \oplus X^{(l)} $ be the sequence of token vectors input to $l\mbox{-}$th layer, where $ X^{(l)}=[x_1^{(l)},...,x_{n}^{(l)}]$ and $n$ is the sequence length of $X^{(l)}$.
We gather the token vectors from $X^{(l)}$ with the top-$k$ informativeness to form the informative sequence $X_{in}^{(l)}$ and the rest vectors to form the uninformative sequence $X_{un}^{(l)}$, where $X_{in}^{(l)} \in  \mathbb{R}^{k \times d_h}$ and $X_{un}^{(l)} \in  \mathbb{R}^{(n\mbox{-}k )\times d_h}$.
The length of the uninformative sequence is reduced by performing certain type of aggregating operations along the sequence dimension, such as \textit{average pooling}:
\begin{equation}
    X_{un}'^{(l)} = \text{Pooling}(X_{un}^{(l)})
\end{equation}
or \textit{weighted average pooling} with informativeness as weights: 
\begin{equation}
\begin{aligned}
    \alpha_{x_{un}^{(l)}} & = \text{softmax}(\text{I}(x^{(l)}_{un})) \\
    X_{un}'^{(l)} & = \text{Pooling}(\alpha X_{un}^{(l)})
\end{aligned}    
\end{equation}
where $x^{(l)}_{un}$ is the token vector in $X_{un}^{(l)}$.  The aggregated sequence $X_{un}'^{(l)} \in  \mathbb{R}^{k' \times d_h}$ and $k'$ is a fixed parameter. After hybrid layer, token sequence is updated to [$x_{cls}^{(l)} \oplus X_{in}^{(l)} \oplus X_{un}'^{(l)}$] and sequence length is shortened by $n\mbox{-}k\mbox{-}k'$. 
Therefore, in addition to the following layers, the computation cost of FFN in $l\mbox{-}$th layer is reduced as well.
It should be noted that the relative position of uninformative tokens should be preserved, which contains their contextual features to a certain extent and they can be captured by aggregating operations.

The parameter $k$ is learnable and progressively shortened. Inspired by ~\citet{powerbert}, we train $n$ learnable parameters to determine the configuration of $k$, denoted $R=[r_1,...,r_n]$. The parameters are constrained to be in the range, i.e., $r_i \in [0;1]$ and added after MHA sub-layer. Given a token vector $x_i$ output by MHA, it is modified by:
\begin{equation}
    x_i \gets r_{pos(x_i)}x_i
\end{equation}
where $pos(x_i)$ is the sorted position of $x_i$ over informativeness. Intuitively, the parameter $r_i$ represents the extent to which the informativeness of the token at $i\mbox{-}$th position is retained. Then, for the $l\mbox{-}$th layer, we obtain the configuration of $k_l$ from the sum of the above parameters, i.e.,
\begin{equation}
\begin{aligned}
    k_l & = \text{ceil}(\text{sum}(l;R))  \\ 
    & s.t. \ \  k_{l+1} \le k_l 
\end{aligned}
\end{equation}

\section{Loss Function}
Let $\Theta$ be the parameters of the baseline BERT model and $\mathcal{L}(\cdot)$ be cross entropy loss or mean-squared error as defined in the original task.
We adopt the multi-task learning idea to jointly minimize the loss in accuracy and total sequence length over all layers. 
\begin{equation}
    \mathop{\text{Loss}}\limits_{\Theta,R} = \mathcal{L}(\Theta,R) + \lambda \sum_{l=1}^{L}l\cdot \text{sum}(l;R)
    \label{loss}
\end{equation}
where L is the number of layers. $\mathcal{L}(\Theta,R)$ controls the accuracy and $\text{sum}(l;R)$  controls the sequence length of $l\mbox{-}$th layer. The hyper-parameter $\lambda$ tunes the trade-off.

The training schema of our model involves three stages, which are given in Algorithm~\ref{process}. 

\begin{algorithm}
\caption{Training Process} 
\label{process}
{\bf Input: } 
{\bf D} = training set \\ 
{\bf Initialize:} $\Theta$ $\leftarrow$ BERT parameters \\
{\bf Initialize:} $R$ $\leftarrow$ uniform distribution 
\begin{algorithmic}[1]
\State fine-tune $\Theta$ on {\bf D} with original loss $\mathcal{L}(\cdot)$
\State add $R$ after MHA sub-layer and fine-tune  $\Theta$ and $R$ with Eq.~\ref{loss}
\State obtain the configuration of $k$ on each layer, then re-train FCA-layer based BERT with $\mathcal{L}(\cdot)$
\end{algorithmic}
\end{algorithm}

\section{Experiments}
\subsection{Datasets}
Our experiments are mainly conducted on GLUE (General Language Understanding Evaluation)~\footnote{https://gluebenchmark.com/} ~\cite{glue} and RACE~\cite{race} datasets. 
GLUE benchmark covers four tasks: Linguistic Acceptability, Sentiment Classification, Natural Language Inference, and Paraphrase Similarity Matching.
RACE is the Machine Reading Comprehension dataset.
\begin{table}
\renewcommand{\arraystretch}{1.5} 
  \centering
  \fontsize{10}{10}\selectfont
    \begin{tabular}{l | c c}
    \hline
    \textbf{Dataset}&\textbf{Task}&\textbf{Input Length} \cr \hline
    \textbf{CoLA}&Acceptability& 64 \cr
    \textbf{RTE}&NLI& 256 \cr 
    \textbf{QQP}&Similarity& 128 \cr 
    \textbf{SST-2}&Sentiment& 64 \cr 
    \textbf{MNLI-m}&NLI&  128\cr 
    \textbf{QNLI}&NLI& 128\cr \hline
    \textbf{RACE} & QA & 512 \cr\hline
 \end{tabular}
  \caption{Statistics of Datasets. }
  \label{Dataset}
\end{table}

For experiments on RACE, we denote the input passage as $P$, the question as $q$, and the four answers as $\{a_1,a_2,a_3,a_4\}$. We concatenate passage, question and each answer as a input sequence $[\text{CLS}] P [\text{SEP}] q [\text{SEP}] a_i [\text{SEP}]$, where $[\text{CLS}]$ and $[\text{SEP}]$ are the special tokens used in the original BERT. The representation of $[\text{CLS}]$ is treated as the single logit value for each $a_i$. Then, a softmax layer is placed on top of these four logits to obtain the normalized probability of each answer, which is used to compute the cross-entropy loss. 

\begin{table*}[h]
\renewcommand{\arraystretch}{1.4} 
  \centering
  \fontsize{10}{10}\selectfont
    \begin{tabular}{|l| c c c c c c c c |}
    \hline
    \textbf{Dataset}&{\textbf{CoLA}}&{\textbf{RTE}}&{\textbf{QQP}}&{\textbf{SST-2}}&{\textbf{MNLI-m}}&{\textbf{QNLI}}&{\textbf{RACE}}& \textbf{Avg.}\cr \hline 
    BERT-base &1.3G & 5.1G & 2.6G& 1.3G  & 2.6G &2.6G&10.2 G & - \cr \hline
    Distil-BERT$_6$ & 0.7G & 2.6G& 1.3G & 0.7G & 1.3G& 1.3G& 5.1 G & - \cr 
    \textbf{Speedup} &2.0x & 2.0x  & 2.0x  & 2.0x & 2.0x & 2.0x & 2.0x & 2.0x  \cr \hline
    FCA-BERT &0.6G & 2.4 G& 1.2G & 0.7G & 1.4G &1.4G &4.4G & - \cr 
    \textbf{Speedup} &2.2x & 2.1x& 2.2x & 1.9x & 1.9x& 1.9x& 2.3x & 2.1x\cr \hline
 \end{tabular}
  \caption{Inference FLOPs. The FLOPs of Distil-BERT$_6$, BERT-PKD$_6$, Tiny-BERT$_6$, Mobile-BERT$_6$ and SNIP are the same and we only list Distil-BERT$_6$'s FLOPs here. The FLOPs of PoWER-BERT is almost the same as that of FCA-BERT as the length of our informative tokens at each layer is set same to the sequence length of PoWER-BERT. The number of coarse units basically does not affect the calculation of FLOPs.}
  \label{speed}
\end{table*}

\begin{table*}
\renewcommand{\arraystretch}{1.4} 
  \centering
  \fontsize{10}{10}\selectfont
    \begin{tabular}{|l| c c c c c c c  c|}
    \hline
    \textbf{Dataset}&{\textbf{CoLA}}&{\textbf{RTE}}&{\textbf{QQP}}&{\textbf{SST-2}}&{\textbf{MNLI-m}}&{\textbf{QNLI}}&{\textbf{RACE}}&{\textbf{Avg.}}\cr \hline 
    BERT-base &55.2 & 67.0& 71.7& 93.0 & 84.8 &91.1& 66.4&75.6  \cr \hline
    Distil-BERT$_6$~\cite{distilbert} &48.8 & 64.2 & 70.2 &89.9 &80.6&88.9&57.9&71.5 \cr 
    BERT-PKD$_6$~\cite{patient} &49.5 & 65.5 & 70.7 &90.4 &81.5 &89.0 &59.3&72.3 \cr
    Tiny-BERT$_6$~\cite{tinybert} &49.2 & 70.2 & 71.1 &91.6 &83.5& 90.5 &59.2&73.6 \cr
    Mobile-BERT$_6$~\cite{mobilebert} &51.1 & \textbf{70.4} & 70.5 &92.6 &\textbf{84.3}& \textbf{91.6} &58.1&74.0 \cr\hline
    FLOP~\cite{flop} & - & - & - & 92.1&- & 89.1& -  & - \cr
    SNIP~\cite{snip} & - & - & - & 91.8&- & 89.5& -  & - \cr
    PoWER-BERT~\cite{powerbert} &51.9 & 65.2& 70.6 &92.2&83.5&89.8&65.3& 74.1 \cr    
    \hline
    FCA-BERT-Pool$_1$ &53.0 & 65.2& 71.1 &92.4&83.5&90.9&65.5&74.5\cr 
    FCA-BERT-Pool$_5$ &54.3 & 66.0& 71.1 &\textbf{93.0}&83.8&90.9&\textbf{66.2}&\textbf{75.0} \cr
    FCA-BERT-Weight$_1$ &\textbf{54.6} & 66.2 & 71.1 &92.6&83.8&90.4&65.8&74.9 \cr 
    FCA-BERT-Weight$_5$ &\textbf{54.6} & 65.6& \textbf{71.4} & \textbf{93.0}&83.9&90.5&66.1&\textbf{75.0} \cr \hline
 \end{tabular}
  \caption{Test results on GLUE and RACE. `Pool' denotes average pooling operation to aggregate uninformative tokens and `Weight' denotes weighted operation. $*_1$ and $*_5$ mean the number of coarse units.}
  \label{result}
\end{table*}

The input length of BERT is set to 512 by default. However, the instances in these datasets are relatively short, rarely reaching 512.
If we keep the default length settings, most of the input tokens are [PAD] tokens. In this way, our model can easily save computational resources by discriminating [PAD] tokens as the uninformative ones, which is meaningless. To avoid this, we constrained the length of the datasets.
The statistic information of the datasets is summarized in Table ~\ref{Dataset}.

\subsection{Evaluation Metrics}
For accuracy evaluation, we adopt Matthew’s Correlation for CoLA, F1-score for QQP, and Accuracy for the rest datasets. For efficiency evaluation, we use the number of floating operations (FLOPs) to measure the inference efficiency, as it is agnostic to the choice of the underlying hardware.

\subsection{Baselines}
We compare our model with both distillation and pruning methods.
Distillation methods contain four models DistilBERT~\cite{distilbert}, BERT-PKD~\cite{patient}, Tiny-BERT~\cite{tinybert}, and Mobile-BERT~\cite{mobilebert}. All four models are distillation from BERT-base and have the same structure (6 transformer layers, 12 attention heads, dimension of the hidden vectors is 768). 
Pruning methods contain FLOP~\cite{flop}, SNIP~\cite{snip}, and PoWER-BERT~\cite{powerbert}. PoWER-BERT~\cite{powerbert} is the state-of-the-art pruning method which reduces sequence length by eliminating word-vectors. To make fair comparisons, we set the length of our informative tokens at each layer same to the sequence length of PoWER-BERT.

\subsection{Implementation Details}
We deploy BERT-base as the standard model in which transformer layers $L\mbox{=}12$, hidden size $d_h\mbox{=}512$, and number of heads $h\mbox{=}12$. All models are trained with 3 epochs. The batch size is selected in list {16,32,64}. The model is optimized using Adam~\cite{adam} with learning rate in range [2e-5,6e-5] for the BERT parameters $\Theta$, [1e-3,3e-3] for $R$. Hyper-parameter ~$\lambda$ that controls the trade-off between accuracy and FLOPs is set in range [1e-3,7e-3].
We conducted experiments with a V100 GPU.
The FLOPs for our model and the baselines were calculated with Tensorflow and batch size=1.
The detailed hyper-parameters setting for each dataset are provided in the Appendix.

\subsection{Main Results}
Table~\ref{result} and Table~\ref{speed} display the accuracy and inference FLOPs of each model on GLUE benchmark respectively. As the FLOPs of PoWER-BERT is almost the same as that of FCA-BERT and the number of coarse units has little affect on FLOPs, Table~\ref{speed} only lists the FLOPs of FCA-BERT.

\noindent \textbf{Comparison to BERT}.
The results demonstrate the high-efficiency of our model, which almost has no performance gap with BERT-base (<\%1 accuracy loss) while reduces the inference FLOPs by half on majority datasets.
Table~\ref{length} presents the sequence length of FCA at each layer, which illustrates substantial reduction of computation length for standard BERT.
For example, the input sequence length for the dataset QQP
is 128. Hence, standard BERT needs to process 128*12=1536 tokens over the twelve layers. In contrast, FCA only tackles
[85, 78, 73, 69, 61, 57, 54, 52, 46, 41, 35, 35] summing to
686 tokens. Consequently, the computational load of  self-attention and the feed forward network is economized significantly.

Among our models, the weighted average pooling operation raises the better performance than the average pooling operation.
The number of coarse units contributes the model accuracy for both two operations, especially for pooling operation. This is reasonable as when the number of coarse units increases, the information stored in each FCA gradually approaches the standard BERT. 
But overmuch coarse units grow FLOPs. 
Therefore, it is necessary to balance impact on FLOPS and performance brought by the coarse units.



\begin{table}
\renewcommand{\arraystretch}{1.5} 
  \centering
  \fontsize{8}{8}\selectfont
    \begin{tabular}{|l| l |}
    \hline
    \textbf{Dataset}&\textbf{Sequence Length} \cr \hline
    \textbf{CoLA} &34, 33, 32, 32, 31, 30, 30, 30, 30, 29, 28, 28  \cr \hline
    \textbf{QQP} &85, 78, 73, 69, 61, 57, 54, 52, 46, 41, 35, 35  \cr \hline
    \textbf{SST-2} &49, 45, 43, 41, 37, 35, 34, 33, 30, 27, 24, 24 \cr \hline
    \textbf{QNLI} &107, 102, 91, 85, 83, 77, 66, 61, 55, 43, 35, 20  \cr \hline
    \textbf{MNLI-m} & 114, 100, 94, 90, 78, 74,66, 62, 51, 40, 28, 24  \cr \hline
    \multirow{2}{*}{\textbf{RTE}} &174, 166, 157, 152, 124, 124,122, 122, 110, 107 \cr
    & 97, 94 \cr \hline
    \multirow{2}{*}{\textbf{RACE}} &261, 244, 230 ,217, 217, 217, 211, 203, 203, 203 \cr
    &  202, 202 \cr \hline
 \end{tabular}
  \caption{Sequence length at each layer.}
  \label{length}
  \vspace{-3mm}
\end{table}

\noindent \textbf{Comparison to Baselines}. We first compare our model to Distil-BERT. 
Our models dramatically outperform Distil-BERT in accuracy by a margin of at least 3 average score. 
As mentioned before, the line of distillation framework is orthogonal to our proposed method. We further investigate whether FCA is compatible with distillation models.
Table~\ref{extension} shows the results of Distil-BERT with FCA-Pool$_5$, which verify that FCA could further accelerate the inference speed on the basis of the distillation model with <1\% loss in accuracy.
As for the SOTA distillation models, Tiny-BERT and Mobile-BERT, our models still outperform them on average performance.
Combined with the results of Table ~\ref{speed} where our models have slightly fewer inference FLOPs than the distillation methods, it can be proved that FCA has better accuracy and computational efficiency than them.


We next compare our model to the SOTA pruning model PoWER-BERT. Their acceleration effects are comparable and we focus on comparing their accuracy.
The results on Table~\ref{result} show that our models achieve better accuracy than PoWER-BERT on all datasets. This is because PoWER-BERT discards the computing units, which inevitably causes information loss. Instead of pruning, FCA layer stockpiles the information of uninformative tokens in a coarse fashion (aggregating operations). 
Moreover, we noticed that coarse units are not always classified as uninformative. In other words, they sometimes participate in the calculation of self-attention as informative tokens. This shows the total informativeness contained in uninformative tokens can not be directly negligible and can be automatically learned  by self-attention.

    

\begin{figure}[tb]
	\centering
	\includegraphics[width=7.6cm]{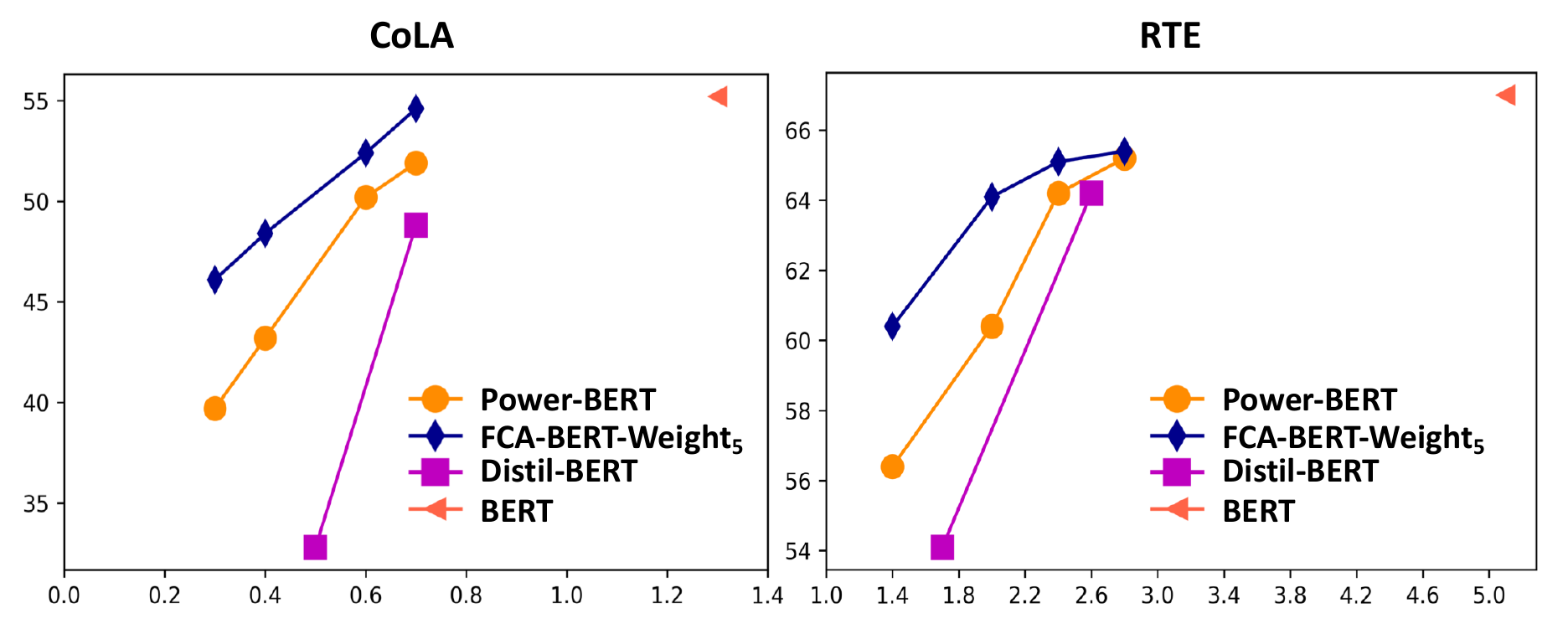}
	\caption{Trade-off between accuracy and FLOPs.}
	\label{trade}
\end{figure}

In order to visually demonstrate the advantages of our model, Figure~\ref{trade} draws curves of trade-off between accuracy and efficiency on three datasets.
The results of FCA-BERT and PoWER-BERT are obtained by tuning the hyper-parameter $\lambda$. 
For DistilBERT, the points correspond to the distillation version with 4 and 6 Transformer layers.
It can be seen that with the decrease of FLOPs, (1) PoWER-BERT and our model outperform Distil-BERT by a large margin; 
(2) our model exhibits the superiority over all the prior methods consistently;
(3) more importantly, the advantage of our model over PoWER-BERT gradually becomes apparent.
This is because PoWER-BERT prunes plenty of computation units to save FLOPs, which results in the dilemma of information loss.
In contrast, our model preserves all information to a certain extent.

\noindent \textbf{Extensions to Other PLMs}.
To explore the generalization capabilities of FCA, we extend FCA to other pre-trained language models (PLMs), such as distil-BERT, BERT-large, and ELECTRA-base~\cite{electra}.  The test results are displayed in Table~\ref{extension}, which proves that FCA is applicable to a variety of models, regardless of model size and variety.

\begin{table}
\renewcommand{\arraystretch}{1.5} 
  \centering
  \fontsize{8}{8}\selectfont
    \begin{tabular}{l| c c c}
    \hline
    \textbf{Dataset}&\textbf{CoLA} &\textbf{RTE} & \textbf{SST-2} \cr \hline
    BERT-large & 60.4 (2.6G) & 70.0 (10.2G) & 94.1 (2.6G) \cr 
    w/ FCA-Pool$_5$ & 59.8 (1.1G) &  66.3 (4.4G)& 93.6 (1.2G) \cr 
    \textbf{Speed-up}& 2.4x & 2.3x & 2.2x \cr \hline
    Distil-BERT$_6$ & 48.8 (0.7G)& 64.2 (2.6G)& 89.9 (0.7G)\cr 
    w/ FCA-Pool$_5$ & 50.2 (0.4G)&  63.9 (1.7G)& 90.4 (0.5G)\cr
    \textbf{Speed-up}& 1.8x &  1.6x & 1.4x \cr \hline
    ELECTRA-base & 62.7 (1.3G) & 75.5 (5.1G)& 95.6 (1.3G)\cr 
    w/ FCA-Pool$_5$ & 62.4 (0.6G)& 75.2 (2.4G) & 95.4 (0.7G)\cr 
    \textbf{Speed-up}& 2.2x &  2.1x & 1.9x \cr \hline

 \end{tabular}
  \caption{Results on other pre-trained language models.}
  \label{extension}
\end{table}




\subsection{Pooling All Tokens}
In this section, we explore that can we not differentiate between tokens and perform the average pooling on all tokens to reduce the computation cost.
To make fair comparisons, we set the length of pooled sequence at each layer equal to the FCA-BERT-Pool$_5$.
The results show that pooling all tokens decreases the model accuracy from 75.0 to 73.8.
This is because the pooling operation weakens the semantic features learned by the informative tokens, which are often decisive for the final prediction. On the contrary, our model does not conduct pooling on informative
tokens and instead delegates the burden of saving computational overhead to uninformative tokens. And this does not cause serious damage to the representative features learned by the model.

\subsection{Distance with Standard BERT}
\begin{figure}
	\centering
	\includegraphics[width=7.5cm]{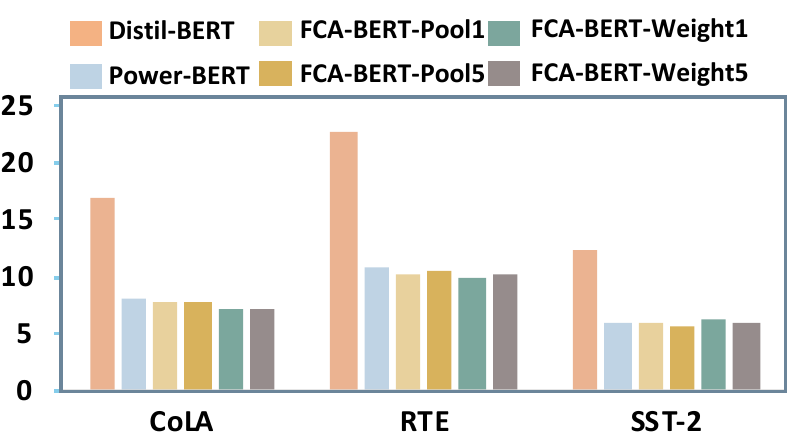}
	\caption{Distance with standard BERT.}
	\label{distance}
\end{figure}

In this section, we further investigate the extent to which these compressed models can retain the essential information of the original BERT.
Concretely, we adopt the Euclidean distance of the CLS representation between BERT and the compressed models as the evaluation metric, which is proportional to the information loss caused by model compression, formally:
$$ \text{Distance(A,B)} = \sum_{k=1}^{M}\sqrt{\sum_{i=1}^{d_h}(\text{A}^{cls}_{i,k}-\text{B}^{cls}_{i,k})}$$
where $M$ is the number of the instances in corresponding dataset.
Table~\ref{distance} shows the distance of baselines and our models with standard BERT.
Combining the results in Table~\ref{result}, it can be found that the distance is consistent with the test accuracy. Large distance leads to low accuracy and vice versa. This provides an inspiration, that is, we can add a distance regulation term to the objective function to forcibly shorten the distance between the compression model and the original BERT, i.e.,
$$ \mathop{\text{Loss}}\limits_{\Theta,R} = \mathcal{L}(\Theta,R) + \lambda \sum_{l=1}^{L}l\cdot \text{sum}(l;R) + \text{Distance}(\cdot)$$

However, the experimental results show that the accuracy has not been significantly improved. This may be because the information learned by the compressed model has reached the limit of approaching the BERT, and the regulation term can not further improve the potential of the compressed model.

\section{Discussion}
Our proposed FCA is dedicated to the classification tasks that only require single-vector representations, and it can not be directly applied to the tasks of requiring to maintain the full input sequence in the output layer, such as NER and extractive MRC. On these tasks, we need to make some modifications of only performing FCA operation over $K$ and $V$ in self-attention and maintaining the full length of $Q$. The Eq.~\ref{attention} is modified to: 
\begin{equation}
\begin{aligned}
    &\text{head}_t = \\ &\text{Attention}(QW_t^Q,\text{FCA}(K)W_t^K,\text{FCA}(V)W_t^V)
\end{aligned}
\end{equation}

We also attempted to maintain the full length of $K$ and $V$ and shorten $Q$, but the experimental results are unsatisfactory.

\section{Conclusion}
In this paper, we propose FCA, a fine- and coarse-granularity hybrid self-attention that reduces the computation cost through progressively shortening the computational sequence length in self-attention.
Experiments on GLUE and RACE datasets show that BERT with FCA achieves 2x reduction in FLOPs over original  BERT with <1\% loss in accuracy.
Meanwhile, FCA offers significantly better trade-off between accuracy and FLOPs compared to prior methods.

\section{Acknowledge}
We would like to thank three anonymous reviewers for their useful feedback.
This work is supported by the National Key Research and Development Program of China under Grant No. 2020AAA0108600.

\bibliography{acl.bib}
\bibliographystyle{acl_natbib}


\end{document}